# Belief functions and default reasoning

S. Benferhat[1]
benferha@irit.fr

A. Saffiotti[2]
asaffio@ulb.ac.be

P. Smets[2]
psmets@ulb.ac.be

[1] IRIT (UPS)
118 Route de Narbonne
Toulouse 31062 Cedex, France

[2] IRIDIA, Université Libre de Bruxelles
50 av. F. Roosevelt
B-1050 Bruxelles, Belgium

**Abstract.** We present a new approach to dealing with default information based on the theory of belief functions. Our semantic structures, inspired by Adams' ε-semantics, are epsilon-belief assignments, where values committed to focal elements are either close to 0 or close to 1. We define two systems based on these structures, and relate them to other non-monotonic systems presented in the literature. We show that our second system correctly addresses the well-known problems of specificity, irrelevance, blocking of inheritance, ambiguity, and redundancy.

## 1. Introduction

Default reasoning is the process of drawing conclusions from i) a set of general rules which may have exceptions, and ii) a set of facts representing the available information (which is often incomplete). The conclusions so drawn are only plausible and can be revised in the light of the new information. The desirable properties for a consequence relation that capture default reasoning have been discussed at length in the AI literature. They can be summarized as follows.

- *Rationality*: the consequence operator used to generate plausible conclusions from a knowledge base should satisfy the rationality postulates proposed by Kraus, Lehmann and Magidor (1990).

- *Specificity*: results obtained from sub-classes should override those obtained from super-classes (Touretzky, 1984). For example, from the knowledge base Δ = {"Birds fly", "Penguins do not fly", "Penguins are birds"}, one should deduce that birds which are penguins do not fly, since penguins are a subclass of birds.

- *Property inheritance:* objects should inherit properties from super-classes unless there is contradiction on that property. For example, from the previous Δ, one should deduce that birds that are red fly, since being red is irrelevant to flying. Also, if we add the rule "Birds have legs", then one should deduce that penguins have legs too, since having legs is not a conflicting property. Failure to perform these deductions is referred to as the problem of "irrelevance" and of "inheritance blocking", respectively.

- *Ambiguity preservation*: in a situation where we have an argument in favor of a proposition, and an independent argument in favor of its negation, we should not conclude anything about that proposition. The most popular example is the so-called Nixon-diamond: knowing that "Quakers are pacifists", "Republicans are not pacifists", and Nixon is both a Quaker and a republican, one should not deduce that Nixon is a pacifist, nor that he is not.[1]

- *Syntax independence*: the consequences of a knowledge base should not depend on the syntactical form used. In particular, they should not be sensitive to duplications of rules in the knowledge base ("redundancy").

In the last decade there have been several proposals for reasoning with default information. Some of them are based on the use of uncertainty models such as probability theory (Adams, 1975; Pearl, 1988), or possibility theory (Dubois and Prade, 1988; Benferhat et al., 1992). Up to now, however, no single system has been reported that correctly addresses all of the desiderata above. In this paper, we show how we can use belief functions, originally developed for modeling uncertainty (Shafer, 1976; Smets, 1988), to build a non-monotonic system that gives a satisfactory answer to all of the above issues.

There have already been a few works on representing default information with belief functions (e.g., Hsia, 1991; Smets and Hsia, 1991). These works require the assessment of numeric values, whose origin is often an open question. Finding a solution free from such assessments would somehow avoid the problem of the origin of the numbers. In this paper, we give another interpretation of default information by using a class of "extreme" belief functions, called epsilon-belief functions, whose non-null masses are either close to 0 or close to 1. The idea of using extreme values is not new to plausible reasoning: for instance, Adams (1975) and Pearl (1990) use extreme probabilities to encode default information; and De Kleer (1990) and Poole (1993) use extreme probabilities for diagnosis.

The rest of this paper is organized as follows. In the next section, we give a short reminder on Adams' ε-semantics and Pearl' System Z, and recall a few notions of the theory of belief functions (see (Shafer, 1976; Smets, 1988) for a complete exposition). In section 3, we introduce ε-belief functions, and show how to use them to define a non-monotonic consequence relation. This relation turns out to be too cautious, and we define two more relations in sections 4 and 5. The first one is based on the least-commitment principle, and is equivalent to system Z. The second one uses Dempster's rule of combination, and is incomparable with the current systems. In section 6, we study this relation in more detail and show that it correctly addresses all of the issues above.

---

[1] Note that this is different from the situation of inconsistency, where we have an argument which supports both a conclusion and its contrary, as in "α→β" and "α→¬β".

## 2. Background

We are interested in *default rules* of the form "generally, if we have $\alpha$ then we have $\beta$", where $\alpha$ and $\beta$ are formulae of some underlying language $\mathcal{L}$. For the goals of this paper, we assume that $\mathcal{L}$ is a classical propositional language. An *interpretation* for $\mathcal{L}$ is an assignment of a truth value in $\{T, F\}$ to each formula of $\mathcal{L}$ in accordance with the classical rules of propositional calculus; we denote by $\Omega$ the set of all such interpretations (also called *worlds*). We say that an interpretation $\omega$ is a *model* of a formula $\alpha$, and write $\omega \models \alpha$ iff $\omega(\alpha) = T$, and denote by $[\alpha]$ the set of all the models of $\alpha$. We write a default rule "generally, if $\alpha$ then $\beta$" as $\alpha \to \beta$, where $\alpha$ and $\beta$ are formulae of $\mathcal{L}$. Note that "$\to$" is a *non-classical* arrow, and it should not be confused with material implication. Given a default rule $d = \alpha \to \beta$, we denote by $\phi_d$ the formula of $\mathcal{L}$ obtained by replacing $\to$ by material implication, namely, $\phi_d = \neg \alpha \vee \beta$. A *default base* is a multiset $\Delta = \{\alpha_i \to \beta_i, i=1,\dots,n\}$ of default rules. We emphasize that a base is a multiset rather than a set, i.e., $\Delta = \{\alpha \to \beta\}$ is different from $\Delta' = \{\alpha \to \beta, \alpha \to \beta\}$.

We use default bases to represent background knowledge about what normally is the case. Given a base $\Delta$, we are interested in defining a consequence relation $\vdash$ between formulae of $\mathcal{L}$ that tells us which consequences we can reasonably drawn from the known facts. We would like $\vdash$ to fulfill the desiderata listed above. For example, given the base $\Delta = \{b \to f, p \to b, p \to \neg f\}$ (where "b" stands for "bird", "f" for "flies", and "p" for "penguin"), we would like to have $b \vdash f$ and $b \wedge p \vdash \neg f$, but not $b \wedge p \vdash f$.

### 2.1. Probabilistic semantics for default rules

Adams (1975), and later Pearl (1988), have suggested a probabilistic interpretation where a default rule $\alpha \to \beta$ is read as a constraint $P(\beta|\alpha) > 1-\varepsilon$, with P a probability distribution and $\varepsilon$ an infinitesimal positive number. Given a set of defaults $\Delta$, they construct a class of probability distributions $A_\varepsilon$ such that, for each distribution P in $A_\varepsilon$ and each default $\alpha \to \beta$ in $\Delta$, $P(\beta|\alpha) > 1-\varepsilon$. A formula $\beta$ is said to be an $\varepsilon$–*consequence* of $\alpha$ with respect to $\Delta$, denoted by $\alpha \vdash_\varepsilon \beta$, if for each $P \in A_\varepsilon$ there exists a real function O such that $\lim_{\varepsilon \to 0} O(\varepsilon) = 0$ and $P(\beta|\alpha) > 1-O(\varepsilon)$. Said differently, $\beta$ is a consequence of $\alpha$ with respect to $\Delta$ if the conditional probability $P(\beta|\alpha)$ is very high provided that the conditional probability of each default in $\Delta$ is very high. Lehmann and Magidor (1992) have shown that $\varepsilon$-consequence is equivalent to the system **P** of Kraus et al (1990), which is commonly regarded as the minimal core of any "reasonable" non-monotonic system.

Adams' $\varepsilon$-consequence is not entirely satisfactory. For example, it suffers from the problem of irrelevance mentioned above: from the default "Generally, birds fly", $\varepsilon$-consequence does not deduce that red birds fly also. To overcome this limitation, Pearl (1990) has proposed a default reasoning system, called Z, based on a ranking of default rules that respects the notion of specificity. Given a default base $\Delta = \{\alpha_i \to \beta_i \mid i = 1,\dots,m\}$, Pearl gives a method to rank-order the rules in $\Delta$ such that the least specific rules (i.e. with most general antecedents) get the least priority. To do this, he defines the notion of *tolerance*: a rule $\alpha \to \beta$ is said to be tolerated by a base $\{\alpha_i \to \beta_i, i = 1,\dots,m\}$ iff $\{\alpha \wedge \beta, \neg\alpha_1 \vee \beta_1, \dots, \neg\alpha_m \vee \beta_m\}$ is consistent. Then, he partitions $\Delta$ into an ordered set $\{\Delta_1, \Delta_2, \dots, \Delta_k\}$ such that rules in $\Delta_i$ are *tolerated* by all rules in $\Delta_i \cup \dots \cup \Delta_k$. From this partition, Pearl induces a ranking $\kappa$ on worlds and, from this, a ranking z on formulae.

Roughly speaking, $\kappa(\omega)$ corresponds to the index of the lowest sub-base that contains a rule violated by $\omega$; and $z(\alpha)$ is the minimum rank of a model of $\alpha$ —so, $z(\alpha)$ can be read as a degree of "abnormality" of $\alpha$ with respect to the rules in $\Delta$. Finally, Pearl defines a non-monotonic inference relation, denoted here by $\vdash_Z$, as follows

$$\alpha \vdash_Z \beta \Leftrightarrow z(\alpha \wedge \beta) < z(\alpha \wedge \neg\beta).$$

An equivalent treatment of default information has been done in the framework of possibility theory (Benferhat et al., 1992).

### 2.2. A reminder on belief functions

Let $\Omega$ be a finite set of worlds, one of them being the actual world. A *basic belief assignment* on $\Omega$ is a function m: $2^\Omega \to [0, 1]$ that satisfies:[2]

$$m(\emptyset) = 0$$

$$\sum_{A \subseteq \Omega} m(A) = 1$$

The term m(A), called the *basic belief mass* given to A, represents the part of a total and finite amount of belief that supports the fact that the actual world belongs to A and does not support the fact that the actual world belongs to a strict subset of A. Any subset A of $\Omega$ for which $m(A) > 0$ is called a *focal element*.

An agent's belief can equivalently be represented by the function bel: $2^\Omega \to [0, 1]$, called a *belief function*, defined by

$$\text{bel}(A) = \sum_{B: B \subseteq A} m(B).$$

The relation between m and bel is one-to-one. The term bel(A) represents the degree of belief, of necessary support, that the actual world belongs to A. Related to bel is another function pl: $2^\Omega \to [0, 1]$, called a *plausibility function*, given by

$$\text{pl}(A) = \sum_{B: B \cap A \neq \emptyset} m(B).$$

The term pl(A) quantifies the degree of plausibility, of potential support, that the agent could give to the fact that the actual world belongs to A. When the focal elements of a basic belief assignment are singletons, then bel = pl is a probability measure. When the focal elements $A_1,\dots,A_n$ are nested (that is, $A_1 \subseteq \dots \subseteq A_n$), bel is called a *consonant belief function*, and bel is a necessity measure and pl is a possibility measure (Zadeh, 1978; Dubois and Prade, 1988). When m has at most one focal element $A \neq \Omega$, it is called a *simple support function*.

When a new piece of evidence telling that the actual world belongs to A becomes available to the agent, his/her belief is revised by the application of the so-called Dempster's rule of conditioning. The basic belief mass m(X) that was supporting the subset X of $\Omega$, now supports $X \cap A$. This transfer of belief masses is described by the following relation, where bel(.|A) denotes the *conditional belief function*.

---

[2] In the transferable belief model (Smets and Kennes, 1994), belief functions are not necessarily normalized, i.e., we can have $m(\emptyset) > 0$. Normalization is assumed here as we only study ratios between bel(B|A) and bel($\Omega$|A), which corresponds to studying the normalized belief functions.

$$bel(X|A) = \frac{bel(X \cup A^c) - bel(A^c)}{bel(\Omega) - bel(A^c)}$$

The impact of a piece of evidence E that bears on $\Omega$ is represented by a belief function bel that describes the agent's beliefs on $\Omega$ given E (and nothing else). Suppose the agent receives two distinct pieces of evidence $E_1$ and $E_2$, and let $bel_1$ and $bel_2$ be the belief functions induced by each evidence individually. The combined effect of $E_1$ and $E_2$ is represented by the belief function $bel_1 \oplus bel_2$ obtained by Dempster's rule of combination. The corresponding basic belief assignment, denoted by $m_1 \oplus m_2$, is given by

$$m_1 \oplus m_2(A) = \frac{\sum_{B \cap C = A} m_1(B) \cdot m_2(C)}{1 - \sum_{B \cap C = \emptyset} m_1(B) \cdot m_2(C)}.$$

## 3. Epsilon-belief functions

In this section, we extend the definition of $\varepsilon$-semantics in a belief function framework. First, we introduce the notion of epsilon-belief functions, whose values are either close to 0 or close to 1. Next, we interpret a default rule $\alpha \to \beta$ as meaning that the conditional belief $bel([\beta]|[\alpha])$ is close to 1. Finally, we define a consequence relation in a natural way: $\alpha \to \beta$ follows from a base of defaults $\Delta$ if, for each epsilon-belief function which satisfies all the default rules in $\Delta$ (i.e., the conditional belief of each rule is close to 1), we have $bel([\beta]|[\alpha])$ close to 1. It turns out that this definition gives us the same results as Adams' $\varepsilon$-consequence relation.

**Definition 1.** An *epsilon-mass assignment* on $\Omega$ is a function $m_\varepsilon : 2^\Omega \to [0,1]$ such that, for each $A \subseteq \Omega$, either $m_\varepsilon(A)=0$, or $m_\varepsilon(A)=\varepsilon_A$, or $m_\varepsilon(A)=1-\varepsilon_A$, where $\varepsilon_A$ is an infinitesimal. The vector $\varepsilon=(\varepsilon_1,...,\varepsilon_k)$ of the infinitesimals appearing in $m_\varepsilon$ is called the *parameter* of m. The belief function induced from $m_\varepsilon$ is called an *epsilon-belief function* ($\varepsilon$-bf).

Throughout this paper, we denote by $bel_\varepsilon$ and $pl_\varepsilon$ the belief and plausibility functions corresponding to a given epsilon-mass assignment $m_\varepsilon$.

**Definition 2.** Let $bel_\varepsilon$ be an $\varepsilon$-bf with parameter $\varepsilon$. We say that $bel_\varepsilon$ is an *ebf-model* of a default rule $\alpha \to \beta$, and write $bel_\varepsilon \models \alpha \to \beta$, iff $\lim_{\varepsilon \to 0} bel_\varepsilon([\beta]|[\alpha]) = 1$, where the limit is taken with respect to all the elements in $\varepsilon$ going to 0.

When working with values that depend on $\varepsilon$, we will often use the notion of one value being infinitely larger than another, written $a >_\infty b$. We say that $a >_\infty b$ if $\lim_{\varepsilon \to 0} b/a = 0$. We also say that a and b are of the same order, written $a \approx b$, if $\lim_{\varepsilon \to 0} b/a = c$, with $c \neq 0$ and finite. The following properties will be useful for working with $\varepsilon$-bf's.[3]

**Lemma 1.** Let $bel_\varepsilon$ be an $\varepsilon$-bf. For any $\alpha, \beta \in \mathcal{L}$,
a) $\lim_{\varepsilon \to 0} bel_\varepsilon([\beta]|[\alpha]) = 1$ iff $pl_\varepsilon([\alpha \wedge \beta]) >_\infty pl_\varepsilon([\alpha \wedge \neg \beta])$.
b) $pl_\varepsilon([\alpha]) \approx \max\{pl_\varepsilon(\omega)|\omega \models \alpha\}$.

---

[3] The proofs of all the results can be found in the long version of this note: (Benferhat et al., 1995).

It is interesting to note that the satisfaction relation $\models$ for $\varepsilon$-bf can also be defined in terms of preferential semantics (Shoham, 1988; Kraus et al., 1990). This will turn out to be useful to relate our systems to other existing ones through well-known results. To create the link, we associate each $\varepsilon$-bf $bel_\varepsilon$ with a preferential order among worlds in $\Omega$ as follows.

**Definition 3.** Let $bel_\varepsilon$ be an $\varepsilon$-bf on $\Omega$, and $\alpha$ a formula of $\mathcal{L}$. The *bel-preference* induced by $bel_\varepsilon$ is the partial order $<_\varepsilon$ given by: $\omega <_\varepsilon \omega'$ iff $pl_\varepsilon(\{\omega'\}) >_\infty pl_\varepsilon(\{\omega\})$. A model $\omega$ of $\alpha$ is called a *bel-preferred model* of $\alpha$ if there is no other world $\omega'$ that satisfies $\alpha$ such that $\omega' <_\varepsilon \omega$.

**Lemma 2.** Let $bel_\varepsilon$ be an $\varepsilon$-bf on $\Omega$. For any $\alpha, \beta$ formulae of $\mathcal{L}$, $bel_\varepsilon \models \alpha \to \beta$ if, and only if, each bel-preferred model of $\alpha$ satisfies $\beta$.

Our next step is to use $\varepsilon$-bf models to define the notion of *entailment* for default bases, i.e., to define which conditional assertions $\alpha \to \beta$ are entailed by a default base $\Delta$. Our first solution is a direct adaptation of the usual definition of logical entailment. We say that an $\varepsilon$-bf $bel_\varepsilon$ is an ebf-model of $\Delta$, written $bel_\varepsilon \models \Delta$, iff $bel_\varepsilon$ is an *ebf-model* of all the rules in $\Delta$. We denote by $Bel_\varepsilon(\Delta)$ the set of all the ebf-models of $\Delta$. Then, a formula $\beta$ is said to be a *bf-consequence* of $\alpha$ (w.r.t. to $\Delta$), denoted by $\alpha \mathrel{|\!\sim}_{bf} \beta$, if and only if each $\varepsilon$-bf $bel_\varepsilon$ of $Bel_\varepsilon(\Delta)$ is an ebf-model of $\alpha \to \beta$, i.e.

**(BF)** $\alpha \mathrel{|\!\sim}_{bf} \beta$ iff for any $bel_\varepsilon$ in $Bel_\varepsilon(\Delta)$, $bel_\varepsilon \models \alpha \to \beta$.

Bf-consequence turns out to be equivalent to the system **P** of Kraus, Lehmann and Magidor (1990), which in turn is equivalent to Adams' $\varepsilon$-system.

**Theorem 1.** For a given $\Delta$, $\alpha \mathrel{|\!\sim}_{bf} \beta$ iff $\alpha \mathrel{|\!\sim}_P \beta$.

The proof, given in (Benferhat et al., 1995), proceeds as follows. Left to right, note that Adams' infinitesimal probability distributions are a special case of our $\varepsilon$-bf, and then $\alpha \mathrel{|\!\sim}_{bf} \beta$ only if $\alpha \mathrel{|\!\sim}_\varepsilon \beta$, only if $\alpha \mathrel{|\!\sim}_P \beta$. Right to left, use lemma 2 to see that each inference relation induced by any $bel_\varepsilon$ in $Bel_\varepsilon(\Delta)$ is preferential. So, $\mathrel{|\!\sim}_{bf}$ satisfies the rules of **P**, and then it contains all preferential consequences of $\Delta$.

This result shows that we can use (infinitesimal) belief functions to give an alternative formalization of the systems $\varepsilon$ and **P**. It also shows that $\mathrel{|\!\sim}_{bf}$ suffers from the same limitations of these systems; in particular, it does not solve the problems of irrelevance and blocking of inheritance. In the next two sections, we propose two ways to define a more bold consequence relation by restricting our attention in **(BF)** to just *some* of the models in $Bel_\varepsilon(\Delta)$.

## 4. Entailment based on least-commitment

One way to select some of the ebf-models of $\Delta$ is by using the notion of being minimally informative: intuitively, we want to look at the consequences of "only knowing" $\Delta$ (and nothing more). A similar approach has been taken, for the case of possibility measures, by Benferhat et al (1992). We recall the following

**Definition 4.** Let $bel_1$ and $bel_2$ two (epsilon-) belief functions over $\Omega$. Then, $bel_1$ is *less committed* that $bel_2$ iff, for any $A \subseteq \Omega$, $pl_1(A) \geq pl_2(A)$.

The *least-commitment principle* (Smets, 1988) states that, in order to model an item of information by a belief function, we should use the least committed belief function that is compatible with the information. Note that the least committed belief function representing a formula $\alpha$ is given by the simple support function that gives mass 1 to $[\alpha]$ and 0 anywhere else. We show how to build an $\varepsilon$bf-model of $\Delta$ based on this principle. We start by allocating a quasi-unitary mass to the set $[\phi_\Delta]$ of the worlds where (the propositional equivalent of) all the defaults in $\Delta$ are satisfied, and the remaining mass $\varepsilon$ to $\Omega$. If there were no conflict in the defaults, this allocation would be an $\varepsilon$bf-model of $\Delta$. When there are conflicts, however, this $\varepsilon$-bf will not satisfy some of the defaults in $\Delta$ — namely, those that inherit a conflicting property from a more general class. Then, we put aside the defaults that are already satisfied, and put almost all of the free mass $\varepsilon$ on the set $[\phi_{\Delta'}]$ corresponding to the still unsatisfied defaults, leaving a small $\varepsilon'$ on $\Omega$. This new $\varepsilon$-bf is an $\varepsilon$bf-model of $\Delta'$ (and of $\Delta$) if we have no conflicts in $\Delta'$. Otherwise, we iterate the procedure until the $\varepsilon$-bf will satisfy all the defaults in $\Delta$.

More precisely, let $\varepsilon = (\varepsilon_1, ..., \varepsilon_n)$ be a vector of infinitesimals such that

($\mathcal{E}_{LC}$)     $\varepsilon_i >_\infty \varepsilon_{i+1}$ for any $i = 1, ..., n-1$.

where n is the cardinality of $\Delta$. We build an $\varepsilon$-bf in the following way.

**Step 0.** Let $i = 0$, $\delta_0 = \Delta$, $sat_0 = \emptyset$, $m_0$ s.t. $m_0(\Omega) = 1$ and $m_0(A) = 0$ otherwise.

**Step 1.** Repeat until $\delta_i = \emptyset$
  1a.   Let $i = i+1$
  1b.   Let $\delta_i = \delta_{i-1} - sat_{i-1}$
  1c.   Let $bel_i$ be the $\varepsilon$-bf given by:
         $m_i(\Omega) = \varepsilon_i$ ; $m_i([\phi_{\delta_i}]) = m_{i-1}(\Omega) - \varepsilon_i$;
         $m_i(A) = m_{i-1}(A)$ otherwise.
  1d.   Let $sat_i = \{d \in \delta_i | bel_i \models d\}$
  1e.   If $sat_i = \emptyset$ and $\delta_i \neq \emptyset$ then Fail.

**Step 2.** Return $bel_{i-1}$.

Note that all the focal elements are nested —the inner one being $[\phi_\Delta]$— and then the final $\varepsilon$-bf returned by Step 2 is a consonant belief function. The procedure fails to find an $\varepsilon$-bf if $\Delta$ is inconsistent; in this case, we have $sat_i = \emptyset$ and $\delta_i \neq \emptyset$.

**Example 1.** Let $\Delta = \{ b \rightarrow f, p \rightarrow b, p \rightarrow \neg f \}$ (where "b" stands for "bird", "f" for "flies", and "p" for "penguin"), and let us apply the previous algorithm. We have $\delta_1 = \Delta$, and $bel_1$ given by: $m_1[\phi_{\delta_1}] = 1-\varepsilon_1$, $m_1(\Omega) = \varepsilon_1$, and $m_1(.) = 0$ elsewhere. We compute the set $sat_1$ of defaults which are satisfied by $bel_1$. We have:

$pl_1([b \wedge f]) = 1-\varepsilon_1+\varepsilon_1 = 1$, $pl_1([b \wedge \neg f]) = \varepsilon_1$,
$pl_1([p \wedge b]) = \varepsilon_1$, $pl_1([p \wedge \neg b]) = \varepsilon_1$,
$pl_1([p \wedge \neg f]) = \varepsilon_1$, $pl_1([p \wedge f]) = \varepsilon_1$.

Therefore, $sat_1 = \{b \rightarrow f\}$. We iterate, and get $\delta_2$ by removing $b \rightarrow f$ from $\delta_1$. Then:

$m_2(\Omega) = \varepsilon_2$; $m_2([\phi_{\delta_2}]) = \varepsilon_1-\varepsilon_2$;
$m_2([\phi_{\delta_1}]) = 1-\varepsilon_1$; $m_2(A) = 0$ otherwise

with $\varepsilon_1$ infinitely larger than $\varepsilon_2$. We have: $pl_2([p \wedge b]) = \varepsilon_1$, $pl_2([p \wedge \neg b]) = \varepsilon_2$, $pl_2([p \wedge \neg f]) = \varepsilon_1$, and $pl_2([p \wedge f]) = \varepsilon_2$. All the defaults in $\delta_2$ are satisfied and the algorithm ends by returning $bel_2$.     ∎

We denote by $Bel_{LC}(\Delta)$ the family of $\varepsilon$-bf's built by the procedure above — the elements of this family differ in the choice of $\varepsilon$, provided that ($\mathcal{E}_{LC}$) is satisfied. This family "behaves well" for our goals: it is a subset of $Bel_\varepsilon(\Delta)$, and it induces a unique ordering $<_\varepsilon$ on the worlds in $\Omega$. The latter property means that we can decide entailment by just looking at one element of $Bel_{LC}(\Delta)$.

**Lemma 3.** Let $\Delta$ be a default base. Then:
a) Any element of $Bel_{LC}(\Delta)$ is an $\varepsilon$bf-model of $\Delta$.
b) Let $bel_1$ and $bel_2$ be two elements of $Bel_{LC}(\Delta)$, and $<_1$ and $<_2$ the corresponding orderings induced on $\Omega$. Then, $<_1 \equiv <_2$.

It is interesting to compare $Bel_{LC}(\Delta)$ with the result of the stratification proposed by Pearl (1988). We can show that the focal elements of the partitions of $Bel_{LC}(\Delta)$ are directly related to the elements of the partition of $\Delta$ obtained by Z.

**Lemma 4.** Let $\Delta = \Delta_1 \cup ... \cup \Delta_n$ be the stratification given by system Z, and let $bel_i$ the $\varepsilon$-bf built by our algorithm at step i. Then, for any default $\alpha \rightarrow \beta$ in $\Delta$,
  a) $\alpha \rightarrow \beta$ is tolerated by $\Delta$ iff $pl_1([\alpha_i]) = 1$
  b) $\alpha \rightarrow \beta \in \Delta_i$ iff $bel_i \models \alpha \rightarrow \beta$.

So, our algorithm produces the same ranking over $\Delta$ than Z. However, our approach does not require an a priori definition of the notion of "tolerance", but relies on the notion of being "less committed". Which of these two notions provides a more natural starting point is a matter of opinion.

We now use the set $Bel_{LC}(\Delta)$ to give our second definition of entailment. It is similar to (**BF**), but we restrict the attention to the $\varepsilon$bf-models that are in $Bel_{LC}(\Delta)$.

(**LC**)    $\alpha \vdash_{lc} \beta$ iff for any $bel_\varepsilon \in Bel_{LC}(\Delta)$, $bel_\varepsilon \models \alpha \rightarrow \beta$.

**Example 2.** We can use the $\varepsilon$-bf $bel_2$ built in Example 1 to check that we have $b \wedge p \vdash_{lc} \neg f$. In fact, we have $pl_2([b \wedge p \wedge \neg f]) = \varepsilon_1$, $pl_2([b \wedge p \wedge f]) = \varepsilon_2$, and $\varepsilon_1 >_\infty \varepsilon_2$. Hence, by lemma 1, $\lim_{\varepsilon \rightarrow 0} bel_2([\neg f]|[b \wedge p]) = 1$, and then $b \wedge p \vdash_{lc} \neg f$.     ∎

As $Bel_{LC}(\Delta)$ is a subset of $Bel_\varepsilon(\Delta)$, lc-consequences include bf-consequences; in particular, they include the consequences of system P. As it turns out, the LC consequence relation is strictly larger than $\vdash_P$, and precisely coincides with Pearl's system Z, as it appears from lemma 4 above.

**Theorem 2.** For a given $\Delta$, $\alpha \vdash_{lc} \beta$ iff $\alpha \vdash_Z \beta$.

## 5. Entailment based on Dempster's rule

The second way that we propose to strengthen the entailment relation (**BF**) is by considering only the $\varepsilon$bf-models of $\Delta$ that can be built by using Dempster's rule of combination. The intuitive argument goes as follows. Suppose we regard each default in $\Delta$ as being one item of evidence provided by one of several *distinct* sources of information.[4] Then, it makes sense to represent each default individually by one belief function, and combine these belief functions by Dempster's rule to obtain a representation of the aggregate effect of all the defaults. We can then define entailment by looking at the conditionals that are satisfied by the combined belief function.

---

[4] The construction given here should extend to the case where each source $S_i$ of information provides an entire base $\Delta_i$ of defaults: we could use the least-commitment principle to build a representative belief function for each $\Delta_i$ as above, and then combine these representatives by Dempster's rule.

There are two technical choices to be made. The first one is which belief function to use to represent each default. Sticking to the arguments used in the last section, we propose to use the least committed εbf that satisfy that default: for a default $d = \alpha \to \beta$, this is the simple support function that allocates a mass of $1-\varepsilon_d$ to $[\phi_d]$, and the remaining mass $\varepsilon_d$ to $\Omega$. We note by $ssf_d$ this function. Given a default base $\Delta = \{d_1,...,d_n\}$, then, we consider the ε-bf obtained by combining all the $ssf_d$'s by Dempster's rule:

$$bel_\oplus = \oplus\{ssf_d \mid d \in \Delta\}.$$

The ε-bf's obtained in this way may be complex: the focal elements are not nested, and the mass values may include products of several $\varepsilon_d$ (or $1-\varepsilon_d$) for different $d$. Luckily, the plausibility of each world can be approximated in a simple way.

**Lemma 5.** Let $bel_\oplus$ be an ε-bf built from $\Delta$ as above. Then, for any world $\omega$ in $\Omega$,

$$pl_\oplus(\{\omega\}) \approx \prod\{\varepsilon_d \mid d \in \Delta \text{ s.t. } \omega \not\models \phi_d\}.$$

If $\omega$ satisfies all formulas of $\Delta$, then $pl_\oplus(\{\omega\})=1$. The second question is how to choose the infinitesimals $\varepsilon_d$'s. We impose constraints on the $\varepsilon_d$'s based on the two following principles:

- *Auto-deduction principle.* We require $bel_\oplus \models \alpha \to \beta$ for each default $\alpha \to \beta$ in $\Delta$ (i.e., $bel_\oplus$ is an εbf-model of $\Delta$).

- *Least commitment principle.* We want each $ssf_d$ to be as un-committed as possible (hence, each $\varepsilon_d$ should be as large as possible).

For each $d = \alpha \to \beta$ in $\Delta$, the first principle generates the constraint (cf. lemma 1)

$$\max\left\{pl_\oplus(\omega) \text{ s.t. } \omega \models \alpha \wedge \beta\right\} >_\infty \max\left\{pl_\oplus(\omega) \text{ s.t. } \omega \models \alpha \wedge \neg\beta\right\}$$

which is equivalent to (cf. Lemma 5)

$$(C_d) \quad \max_{\omega \models \alpha \wedge \beta}\left\{\prod_{d: \omega \not\models \phi_d} \varepsilon_d\right\} >_\infty \max_{\omega \models \alpha \wedge \neg\beta}\left\{\prod_{d: \omega \not\models \phi_d} \varepsilon_d\right\}$$

By solving the system given by all the $C_d$'s, we can get constraints of the form $\varepsilon_i >_\infty \varepsilon_j$ between some of the elements of ε (and between products of elements). The second principle can be used to sanction equivalencies between unconstrained elements of ε by the following argument. Suppose that, for some $\varepsilon_i$ and $\varepsilon_j$, neither $\varepsilon_i >_\infty \varepsilon_j$ nor $\varepsilon_j >_\infty \varepsilon_i$ is the case; as both $\varepsilon_i$ and $\varepsilon_j$ should be made as large as possible, no one can be larger than the other, and then $\varepsilon_i = \varepsilon_j$.

We now describe an algorithm to solve a set $C$ of constraints of the form $(C_d)$. We call *term* any product of elements of ε, and *complex term* a term containing at least two elements of ε. The algorithm returns a set $\xi = \{\xi_0,...,\xi_n\}$ of equivalence classes of terms such that: 1) all the terms in a class are of the same order; 2) any term in $\xi_i$ is infinitely larger than any term in $\xi_j$ when $i < j$; and 3) any element of ε is in some class.

**Step 0.** Let $i = 0$, $A_t=\{t_i \mid t_i$ is a term and $1>_\infty t_i$ is in $C\}$.
    0.a    Let $\xi_0$ be a set of $\varepsilon_i$ such that there exists a complex term t in $A_t$ which contains $\varepsilon_i$
    0.b.    If $\xi_0=\emptyset$ then $\xi_0=A_t$; else $\xi_1=A_t-\xi_0$ and $i=1$
**Step 1.** Repeat until $C = \emptyset$
    1a.    Let $i=i+1$. Remove from $C$ any satisfied constraint
    1b.    Let $A_t$ be a set of terms in $C$ which does not appear in the right side of any constraint of $C$
    1c.    Let $\xi_i$ be a set of $\varepsilon_i$ which does not appear in any $\xi_{j<i}$ and where there exists a complex term t in $A_t$ which contains $\varepsilon_i$
    1.d.    If $\xi_i \neq \emptyset$ then $A_t := A_t - \xi_i$ and $i = i+1$
    1.e.    Let $A_\varepsilon$ be the set of $\varepsilon_i$ which does not appear neither in any constraint of $C$ nor in any $\xi_{j<i}$.
    1.f.    Let $\xi_i = A_t + A_\varepsilon$
**Step 2.** Return the sets $\xi_{j=1,i}$.

**Example 3.** Let again $\Delta = \{b \to f, p \to \neg f, p \to b\}$. The simple support functions corresponding to the three defaults in $\Delta$ are given by

$m_1([\neg b \vee f]) = 1 - \varepsilon_1$, $m_1(\Omega) = \varepsilon_1$,
$m_2([\neg p \vee \neg f]) = 1 - \varepsilon_2$, $m_2(\Omega) = \varepsilon_2$, and
$m_3([\neg p \vee b]) = 1 - \varepsilon_3$, $m_3(\Omega) = \varepsilon_3$.

The requirement of auto-deductivity gives us the following three constraints:

$\max\{pl_\oplus(\{\omega\}) \mid \omega \models b \wedge f\} >_\infty \max\{pl_\oplus(\{\omega\}) \mid \omega \models b \wedge \neg f\}$
i.e., $1 >_\infty \varepsilon_1$

$\max\{pl_\oplus(\{\omega\}) \mid \omega \models p \wedge \neg f\} >_\infty \max\{pl_\oplus(\{\omega\}) \mid \omega \models p \wedge f\}$
i.e., $\max\{\varepsilon_1, \varepsilon_3\} >_\infty \varepsilon_2$

$\max\{pl_\oplus(\{\omega\}) \mid \omega \models p \wedge b\} >_\infty \max\{pl_\oplus(\{\omega\}) \mid \omega \models p \wedge \neg b\}$
i.e., $\max\{\varepsilon_1, \varepsilon_2\} >_\infty \varepsilon_3$

Let us apply the previous algorithm. The set $\xi_0$ contains exactly one element $\varepsilon_1$. When we put $\varepsilon_0$ in the highest value then all the constraints will be satisfied. Therefore the result is $\xi_0=\{\varepsilon_1\}$, $\xi_1=\{\varepsilon_2, \varepsilon_3\}$, and $\varepsilon_1 >_\infty \varepsilon_2 = \varepsilon_3$. ∎

We denote by $Bel_\oplus(\Delta)$ the family of ε-bf's built by Dempster's rule and whose parameter ε satisfy the constraints above. As we did for the $Bel_{LC}(\Delta)$ family, we make sure that the elements of $Bel_\oplus(\Delta)$ have the right properties for our goals: they are εbf-models of $\Delta$, and they induce a unique ordering on $\Omega$.

**Lemma 6.** Let $\Delta$ be a default base. Then:
a) Any element of $Bel_\oplus(\Delta)$ is an εbf-model of $\Delta$.
b) Let $bel_1$ and $bel_2$ be elements of $Bel_\oplus(\Delta)$, and $\prec_1$ and $\prec_2$ the corresponding orderings induced on $\Omega$. Then, $\prec_1 \equiv \prec_2$.

Our third and last definition of entailment, called LCD (Least-Commitment plus Dempster's rule), is obtained by focusing on the εbf-models that are in $Bel_\oplus(\Delta)$.

**(LCD)** $\alpha \mathrel{\vdash}_{lcd} \beta$ iff for any $bel_\varepsilon$ in $Bel_\oplus(\Delta)$, $bel_\varepsilon \models \alpha \to \beta$

Note that, as all the elements of $Bel_\oplus(\Delta)$ are εbf-model of $\Delta$, the $\mathrel{\vdash}_{lcd}$ relation is as least as strong as $\mathrel{\vdash}_{bf}$. In particular, $\mathrel{\vdash}_{lcd}$ satisfies the KLM properties for system **P** (Kraus et al., 1990). In fact, **LCD** is strictly stronger than **P**. For example, **LCD** correctly addresses the irrelevance problem, as shown by the following example.

**Example 4.** We first show how to use the result of the previous example to verify that $b \wedge p \mathrel{\vdash}_{lcd} \neg f$. In fact, by applying lemmas 1(b) and 5, we have $pl_\oplus([b \wedge p \wedge f]) \approx \varepsilon_2$ and $pl_\oplus([b \wedge p \wedge \neg f]) \approx \varepsilon_1$, and we know that $\varepsilon_1 >_\infty \varepsilon_2$. Next, con-sider a new property "red" (r) unrelated to b, p and f. We expect that red birds fly (note that this is not the case in system P). For any bel in $Bel_\oplus$, and its corresponding pl, we have

$$pl([b \wedge r \wedge f]) = \max\{pl([b \wedge r \wedge f \wedge p]), pl([b \wedge r \wedge f \wedge \neg p])\}$$
$$= \max\{\varepsilon_2, 1\} = 1$$
$$pl([b \wedge r \wedge \neg f]) = \max\{pl([b \wedge r \wedge \neg f \wedge p]), pl([b \wedge r \wedge \neg f \wedge \neg p])\}$$
$$= \max\{\varepsilon_1, \varepsilon_1\} = \varepsilon_1$$

Hence, we have $pl([b \wedge r \wedge f]) >_\infty pl([b \wedge r \wedge \neg f])$, which implies $b \wedge r \mathrel{\mid\!\sim}_{lcd} f$, as was desired. ∎

The following theorem summarizes the relation between (LCD) and system **P**.

**Theorem 3.** For a given $\Delta$, if $\alpha \mathrel{\mid\!\sim}_P \beta$ then $\alpha \mathrel{\mid\!\sim}_{lcd} \beta$. The converse is not true.

## 6  Analysis of LCD-consequence

We have seen that the LCD consequence relation gives us strictly more than system **P**; in particular, it correctly answers the problem of irrelevance. In this section, we study in more detail the patterns of reasoning that are captured by LCD. To do this, we consider the desiderata listed in the introduction, and show how LCD addresses them. We also contrast the LCD solution with the one obtained by other existing systems that go beyond system **P**.

We start by property inheritance. Several systems, including Pearl's system **Z**, suffer from the problem of inheritance blocking. The canonical example is built by adding to the usual penguin problem the default $b \to l$ (generally, birds have legs). From this, system **Z** cannot deduce that penguins have legs also, i.e., $p \mathrel{\not\mid\!\sim}_Z l$.[5] By contrast, LCD allows that deduction, as shown below.

**Example 5.** Let $\Delta = \{b \to f, p \to \neg f, p \to b, b \to l\}$, where l stands for legs. The simple support functions are those in Example 3, plus

$$m_4([\neg p \vee l]) = 1 - \varepsilon_4, \quad m_4(\Omega) = \varepsilon_4.$$

The constraint that $pl_\oplus$ must satisfy are the same as in Example 3, plus

$$\max\{pl_\oplus(\{\omega\})|\omega \models b \wedge l\} >_\infty \max\{pl_\oplus(\{\omega\})|\omega \models b \wedge \neg l\}$$
i.e., $1 >_\infty \varepsilon_4$

We apply our algorithm to this set. The first layer, $\xi_0$, contains exactly two elements: $\varepsilon_1$ and $\varepsilon_4$. Once we constrain $\varepsilon_1$ and $\varepsilon_4$ to have the highest value, all the constraints are satisfied. Therefore we get: $\xi_0 = \{\varepsilon_1, \varepsilon_4\} >_\infty \xi_1 = \{\varepsilon_2, \varepsilon_3\}$. To see if penguins have legs, we compute

$$pl([p \wedge l]) = \max\{pl([p \wedge l \wedge b \wedge f]), pl([p \wedge l \wedge b \wedge \neg f]),$$
$$pl([p \wedge l \wedge \neg b \wedge f]), pl([p \wedge l \wedge \neg b \wedge \neg f])\}$$
$$= \max\{\varepsilon_2, \varepsilon_1, \varepsilon_2 \varepsilon_3, \varepsilon_2\} = \varepsilon_1$$
$$pl([p \wedge \neg l]) = \max\{pl([p \wedge \neg l \wedge b \wedge f]), pl([p \wedge \neg l \wedge b \wedge \neg f]),$$
$$pl([p \wedge \neg l \wedge \neg b \wedge f]), pl([p \wedge \neg l \wedge \neg b \wedge \neg f])\}$$
$$= \max\{\varepsilon_2 \varepsilon_4, \varepsilon_1 \varepsilon_4, \varepsilon_2 \varepsilon_3 \varepsilon_4, \varepsilon_2 \varepsilon_4\} = \varepsilon_1 \varepsilon_4$$

Therefore, $pl([p \wedge l]) >_\varepsilon pl([p \wedge \neg l])$, which implies $p \mathrel{\mid\!\sim}_{lcd} l$ as was desired. ∎

---

[5] Goldszmidt and Pearl (1991) have suggested an extension of **Z**, called **Z**[+], that correctly handles this example. Unfortunately, **Z**[+] does not solve the problem of inheritance blocking in general: if we add the rules "Generally, legless birds are birds" and "Generally, legless birds do not have legs" to our base, then **Z**[+] cannot deduce both of "Legless birds fly" and "Penguins have legs"—it will just deduce one of them, depending on the ranking. This problem is solved by LCD.

Another desiderata listed in the introduction was the ability to stay uncommitted in cases of ambiguity. The following example shows a case of ambiguity where system **Z** would deduce an undesired result, while LCD does not.

**Example 6.** Let $\Delta = \{b \to f, p \to \neg f, p \to b, m \to f\}$, where the last default means "Generally, objects with metal-wings fly." The simple support functions are again those in Example 3, plus the following one:

$$m_4([\neg m \vee f]) = 1 - \varepsilon_4, \quad m_4(\Omega) = \varepsilon_4.$$

The constraints that $pl_\oplus$ must satisfy are the same as in Example 3, plus

$$\max\{pl_\oplus(\{\omega\})|\omega \models m \wedge f\} >_\infty \max\{pl_\oplus(\{\omega\})|\omega \models m \wedge \neg f\}$$
i.e., $1 >_\infty \varepsilon_4$

We get the same ordering as before, with $\varepsilon_4$ in the top class: $\xi_0 = \{\varepsilon_1, \varepsilon_4\} >_\infty \xi_1 = \{\varepsilon_2, \varepsilon_3\}$. Consider now a bird that is a penguin and has metal wing. Given the base $\Delta$, we can not say whether or not this beast will fly — we are in a case of ambiguity. And indeed we have:

$$pl([b \wedge p \wedge m \wedge f]) = \varepsilon_2$$
$$pl([b \wedge p \wedge m \wedge \neg f]) = \varepsilon_1 \varepsilon_4.$$

As the ordering above says nothing of the relative magnitude of $\varepsilon_2$ and $\varepsilon_1 \varepsilon_4$, we do not have neither $b \wedge p \wedge m \mathrel{\mid\!\sim}_{lcd} f$ nor $b \wedge p \wedge m \mathrel{\mid\!\sim}_{lcd} \neg f$. Notice, by contrast, that **Z** would give us the arbitrary result $b \wedge p \wedge m \mathrel{\mid\!\sim}_Z \neg f$. ∎

The following theorem summarizes the relation between LCD and system **Z**.

**Theorem 4.** The consequence relations $\mathrel{\mid\!\sim}_{lcd}$ and $\mathrel{\mid\!\sim}_Z$ are incomparable.

The last desiderata in the introduction was syntax-independence. The following example shows that LCD is not sensitive to duplications of rules in the default base.

**Example 7.** Consider a variant of the Quaker-Republican problem where the rule "Generally, Quaker are pacifists" has been duplicated: $\Delta = \{q \to p, q \to p, r \to \neg p\}$. By using a lexicographic approach, we would deduce $q \wedge r \to p$, while we would prefer to acknowledge the ambiguity and deduce nothing. In LCD, we have

$$m_1([\neg q \vee p]) = 1 - \varepsilon_1, \quad m_1(\Omega) = \varepsilon_1,$$
$$m_2([\neg q \vee p]) = 1 - \varepsilon_2, \quad m_2(\Omega) = \varepsilon_2,$$
$$m_3([\neg r \vee \neg p]) = 1 - \varepsilon_3, \quad m_3(\Omega) = \varepsilon_3,$$

together with the constraints

$$1 >_\infty \varepsilon_1 \varepsilon_2, \quad 1 >_\infty \varepsilon_1 \varepsilon_2, \quad 1 >_\infty \varepsilon_3$$

By applying our algorithm, we get one single class, and so $\varepsilon_3 = \varepsilon_1 \varepsilon_2$. Then

$$pl([q \wedge r \wedge p]) = \varepsilon_3 \text{ and } pl([q \wedge r \wedge \neg p]) = \varepsilon_1 \varepsilon_2,$$

and hence, as desired, we have neither $q \wedge r \mathrel{\mid\!\sim}_{lcd} p$ nor $q \wedge r \mathrel{\mid\!\sim}_{lcd} \neg p$. ∎

We have shown that the LCD consequence relation behaves well with respect to all of the desiderata stated in the introduction. Unfortunately, there are cases where LCD gives us results whose intuitive acceptability is questionable.

**Example 8.** Consider the Quaker-Republican problem with the extra rule "Generally, ecologists are pacifist": $\Delta = \{q \to p, e \to p, r \to \neg p\}$. We have

$$m_1([\neg q \vee p]) = 1 - \varepsilon_1, \quad m_1(\Omega) = \varepsilon_1,$$
$$m_2([\neg e \vee p]) = 1 - \varepsilon_2, \quad m_2(\Omega) = \varepsilon_2,$$
$$m_3([\neg r \vee \neg p]) = 1 - \varepsilon_3, \quad m_3(\Omega) = \varepsilon_3,$$

together with the constraints

$1 >_\infty \varepsilon_1$, $1 >_\infty \varepsilon_2$, $1 >_\infty \varepsilon_3$

Since all the elements of $\varepsilon$ are free, our algorithm puts all of them in the same class, i.e., $\varepsilon_1 = \varepsilon_2 = \varepsilon_3$. Then, we have

$pl([q \wedge e \wedge r \wedge p]) \approx \varepsilon_3 >_\infty pl([q \wedge e \wedge r \wedge \neg p]) \approx \varepsilon_1 \varepsilon_2$,

and so ecologists who are Quakers and republicans are pacifist: $q \wedge e \wedge r \mathrel{\vert\!\sim}_{lcd} p$. ∎

The last example may be disappointing, in that we may consider that we are in a case of ambiguity and we should stay silent. The reason for the answer given by LCD is to be found in the multi-source interpretation of the approach: having two sources to independently justify a conclusion is regarded as a stronger reason to accept that conclusion. The assumption of independence between the sources is essential to use Dempster's rule of combination; it may be interesting to study variants of LCD based on different rules of combination.

## 7. Three special cases

The LCD consequence relation has been built by giving each default $d$ an infinitesimal weight $\varepsilon_d$, and combining these weights by Dempster's rule of combination. We then used the auto-deduction and least-commitment principles to constrain the possible values of the $\varepsilon_d$'s. As an alternative, we could have imposed some *a-priori* relation between the $\varepsilon_d$'s. In this section, we show three special cases of consequence relations obtained in the second way. The relations we get are already known in the literature: in the first case, we get the so-called "penalty logic" (Pinkas, 1992; Dupin et al., 1994); in the second case, we get the lexicographic approach (Dubois et al., 1992; Benferhat et al., 1993; Lehmann, 1993); and the last case leads to the preferred sub-theories of Brewka (1989).

Consider a default base $\Delta$, and let $\{\Delta_0, \Delta_1, ..., \Delta_n\}$ be the partition of $\Delta$ given by System Z. As we did for LCD, we associate each default $d$ in $\Delta$ with the following simple support function

$m_d([\phi_d]) = 1 - \varepsilon_d$, $m_d(\Omega) = \varepsilon_d$, $m_d(\text{elsewhere}) = 0$.

The three special cases we want to analyze are obtained by fixing the value of the $\varepsilon_d$'s infinitesimals as follows.

For the fist case, we choose one "global" infinitesimal $\varepsilon$, and we let, for each $d \in \Delta_i$, $\varepsilon_d = \varepsilon^i$. Then by applying Dempster's rule of combination we get for each interpretation:

$$pl_\oplus(\{\omega\}) \approx \prod \{\varepsilon_d \mid d \in \Delta \text{ s.t. } \omega \ne \phi_d\}$$
$$\approx \prod_{i=1,n} \varepsilon^{(i \cdot k_i(\omega))}$$
$$= \varepsilon^{[\sum_{i=1,n} (i \cdot k_i(\omega))]},$$

where $k_i(\omega)$ is the number of defaults of $\Delta_i$ which are not satisfied by the interpretation $\omega$. We can interpret the resulting mass assignment $m_\oplus$ in terms of costs: each piece of information $d$ of $\Delta_i$ is associated to the cost $c(d) = i$, interpreted as the price to pay if $d$ is not satisfied. Then each interpretation $\omega$ is associated the sum of the costs of pieces of information of $\Delta$ which are falsified by $\omega$, namely:

$C(\omega) = \sum_{\omega \ne \phi_d, d \in \Delta} c(d)$

We can show that

$C(\omega) < C(\omega')$ iff $Pl_\oplus(\omega) >_\infty Pl_\oplus(\omega')$.

(Obvious, since $pl_\oplus(\omega) \approx \varepsilon^{\sum_{i=1,n} [i \cdot k_i(\omega)]} = \varepsilon^{C(E)}$.) Hence, the non-monotonic consequence relation obtained by using Dempster's combination and the $\varepsilon_d$'s defined above generates the same results as the inference relation based on penalty logic (Pinkas, 1991; Dupin et al., 1994).

For the second case, we equate all the $\varepsilon_d$'s of the defaults $d$ that belong to the same layer $\Delta_i$. More precisely, we associate an infinitesimal $\varepsilon_i$ to each layer $\Delta_i$, and, for each $d \in \Delta_i$, let $\varepsilon_d = \varepsilon_i$. We then ask that $\varepsilon_1$ is a positive real number infinitely small, and that, for any $i > 1$,

$$\prod_{j=1}^{i-1} \varepsilon_j^{|\Delta_j|} >_\infty \varepsilon_i \,,$$

where $|\Delta_j|$ is number of default rules in the stratum $\Delta_j$. We can show that the order on different interpretations obtained by using these $\varepsilon_d$'s can be characterized as follows, where $[\omega]_i$ is the number of defaults of $\Delta_i$ satisfied by $\omega$.

**Proposition 1.** $pl(\omega) >_\infty pl(\omega')$ if and only if there exist a positive number $1 \le i < n$ such that:
- $\forall j > i, |[\omega]_j| = |[\omega']_j|$, and
- $|[\omega]_i| > |[\omega']_i|$

This ordering corresponds to the so-called *Lexicographical ordering* defined in (Dubois et al., 1992; Benferhat et al., 1993; Lehmann, 1993), and hence the consequence relation obtained by using Dempster's combination and the $\varepsilon_d$'s defined above is the same as the one obtained by the lexicographic approaches. This ordering has also been considered in diagnosis by De Kleer (1990) and Lang (1994).

For the last case, let $d_{ij}$ be the j-th default (according to some arbitrary enumeration) in the i-th layer $\Delta_i$. Then, we associate $d_{ij}$ to an infinitesimal $\varepsilon_{ij}$ such that:

for a given $i$, $\varepsilon_{ij}$ and $\varepsilon_{ik}$ for $k \ne j$ are incomparable,
$\varepsilon_{1j}$ are positive real numbers infinitely small, and
$\prod_{j=1,i-1} \prod_{k=1,|\Delta_j|} \varepsilon_{jk} >_\infty \varepsilon_{il}$ for $l=1,|\Delta_i|$, and $i > 1$

where $|\Delta_x|$ is number of default rules in layer $\Delta_x$. Then, we can show that the order on the interpretations can be characterized in the following way, where $[\omega]_i$ is the number of defaults of $\Delta_i$ satisfied by $\omega$.

**Proposition 2.** $pl(\{\omega\}) >_\infty pl(\{\omega'\})$ if and only if there exist a positive number $1 \le i < n$ such that:
- $\forall j > i, [\omega]_j = [\omega']_j$, and
- $[\omega]_i \supseteq [\omega']_i$ (and not $[\omega']_i \supseteq [\omega]_i$).

This ordering corresponds to the *preferred sub-theories* originally proposed by Brewka (1989), and later independently introduced in (Dubois et al., 1992) in the setting of possibilistic logic.[6] Hence, the consequence relation obtained by using Dempster's combination and the $\varepsilon_d$'s defined above is the same as the one obtained by the system of Brewka.

As it is the case for our LCD system, all the three particular cases discussed in this section are strictly stronger than system P. In general, we state the following:

**Theorem 5.** LCD-consequence is incomparable with all of: penalty logic; the lexicographic approaches; and the preferred sub-theories approach.

---

[6] Brewka's preferred subtheories have also been used by Boutilier (1992) in system Z to define a nonmonotonic inference relation, and by Baral (1992) to combine belief bases.

## 8. Conclusions

We have detailed a new approach to deal with default information based on a special class of belief functions, and have used it to define three non-monotonic consequence relations. The last one, LCD-consequence, appears to be particularly attractive. We have proved that LCD is stronger than system **P**, and thus it satisfies the rationality postulates of Kraus, Lehmann and Magidor (1990). Moreover, we have given examples showing that LCD correctly addresses the well-known problems of irrelevance (example 4); of blocking of inheritance (example 5); of ambiguity (example 6); and of redundancy (example 7). In this, LCD has a distinctive advantage over all currently existing approaches. Finally, we have shown that the construction used to define LCD can be used to build alternative definitions of several existing systems of default reasoning, including penalty logic, the lexicographic approach, and Brewka's preferred sub-theories. Hence, our belief-function based semantics for defaults appears to be able to capture many of the current systems in a common framework.

It is interesting to notice that, despite its good behaviour, LCD does not satisfy rational monotonicity. We are currently studying different rules that can characterize LCD-consequence. Moreover, although LCD is insensitive to the number of repetitions of the same default rule, it is sensitive to the number of different rules supporting the same conclusion. This is not surprising given that LCD is based on the interpretation of defaults as items of information provided by independent sources. We plan to explore variants of LCD that abandon the assumption of independence.

**Acknowledgements.** Work by the second and third author has been partially supported by the ARC "BELON", founded by the Communauté Française de Belgique.

## References

Adams, E. W. (1975) *The logic of conditionals*. Reidel, Dordrecht.

Baral, S. Kraus, J. Minker, V.S. Subrahmanian (1992) Combining knowledge bases consisting in first order theories. Computational Intelligence, 8(1), 45-71.

Benferhat, S., Dubois, D. and Prade, H. (1992) Representing default rules in possibilistic logic. Procs. of the 3rd Conf. on Knowledge Representation and Reasoning KR'92.

Benferhat, S., Cayrol, C., Dubois, D., Lang, J. and Prade, H. (1993) Inconsistency management and prioritized syntax-based entailment. Procs. of IJCAI'93.

Benferhat, S., Saffiotti, A. and Smets, Ph (1995) Belief functions and default reasoning. Tech. Rep. TR/IRIDIA/95-5, Université Libre de Bruxelles, Belgium. Available on: ftp://iridia.ulb.ac.be/pub/saffiotti/uncertainty/tr_95_5.ps.

Boutilier, G. (1992). What is a Default priority?. *Proc. of the 9th Canadian Conf. on AI (AI'92)*, Vancouver, May 1992, 140-147.

Brewka, G. (1989) Preferred subtheories: an extended logical framework for default reasoning. Proc. of the 11th Inter. Joint Conf. on AI (IJCAI'89), Detroit, 1043-1048.

De Kleer, J. (1990) Using Crude Probability Estimates to Guide Diagnosis. *Artificial Intelligence*, 45, 381-391.

Dubois, D., Lang, J. and Prade, H. (1992) Inconsistency in possibilistic knowledge bases — To live or not live with it. In: Fuzzy Logic for the Management of Uncertainty (L.A. Zadeh, J. Kacprzyk, eds.), Wiley, 335-351.

Dubois, D. and Prade, H. (1988) (with the collaboration of Farreny H., Martin- Clouaire R., Testemale C.) *Possibility Theory – An Approach to Computerized Processing of Uncertainty*. Plenum Press, New York

Dupin de Saint Cyr, F., Lang, J., and Schiex, N. (1994) Penalty logic and its link with Dempster-Shafer theory. Proc. of UAI'94, 204-211.

Gabbay, D. (1985) Theoretical foundations for non-monotonic reasoning in expert systems. In: K.R. Apt (Ed.) *Logics and models of Concurrent Systems*. Springer Verlag, Berlin, 439-457.

Goldszmidt, M. and Pearl, J. (1991) System $Z^+$: A formalism for reasoning with variable-strength defaults. Procs. of the AAAI Conference, Anaheim, CA.

Hsia, Y-T. (1991) A belief-Function Semantics for Cautious Non-Monotonicity, Technical Report TR/IRIDIA/91-3, Université Libre de Bruxelles, Belgium.

Kraus, S., Lehmann, D. and Magidor, M. (1990) Non-monotonic reasoning, preferential models and cumulative logics. *Artificial Intelligence* 44, 167-207.

Lang, J. (1994) Syntax-based default reasoning as probabilistic model-based diagnosis. In Proc. of UAI'94, 391-398.

Lehmann, D. (1993) Another perspective on Default Reasoning. Technical report, Hebrew University.

Lehmann, D. and Magidor, M. (1992) What does a conditional knowledge base entail? *Artificial Intelligence*, 55, 1-60.

Pearl, J. (1988) *Probabilistic Reasoning in Intelligent Systems: Networks of Plausible Inference*. Morgan Kaufmann, CA.

Pearl J. (1990) System Z: A natural ordering of defaults with tractable applications to default reasoning. Procs. of the Conf. on Theoretical Aspects of Reasoning about Knowledge, 121-135.

Pinkas, G. (1991). Propositional nonmonotonic reasoning and inconsistency in symmetric neural networks. Procs. of the 12th IJCAI, 525-530, Sydney, Australia.

Poole, D. (1993) Average-case analysis of a search algorithm for estimating prior and posterior probabilities in Bayesian networks with extreme probabilities. Procs. of IJCAI'93, 606-612.

Shafer, G. (1976) *A Mathematical Theory of Evidence*. Princeton University Press, New Jersey.

Shoham, Y. (1988) *Reasoning About Change — Time and causation from the standpoint of Artificial Intelligence*. MIT Press, Cambridge, MA.

Smets, Ph. (1988) Belief functions. In: Ph. Smets, E. H. Mamdani, D. Dubois, H. Prade (Eds.) *Non-Standard Logics for Automated Reasoning*. Academic Press, New York, 253-286.

Smets, Ph and Hsia, Y-T. (1991) Default Reasoning and the Transferable Belief Model. In: P. Bonissone, M. Henrion, L. Kanal and J. Lemmer (Eds.) *Uncertainty in Artificial Intelligence 6*. North Holland, Amsterdam, 495-504.

Smets, Ph. and Kennes, R. (1994) The transferable belief model. *Artificial Intelligence* 66, 191-234.

Touretzky, D. (1984) Implicit ordering of defaults in inheritance systems. Procs. of AAAI'84, 322-325.

Zadeh, L.A. (1978) Fuzzy sets as a basis for a theory of possibility. *Fuzzy Sets and Systems* 1, 3-28.